\pdfoutput=1
\documentclass[11pt]{article}

\newcommand{\intra}{\emph{in-item} representation}
\newcommand{\inter}{\emph{cross-item} interactions}
\newcommand{\neglect}{\emph{item negligence}}

\newcommand{\appref}[1]{Appendix~\ref{#1}}
\newcommand{\secref}[1]{§\ref{sec:#1}}

\usepackage[final]{acl}

\usepackage{times}
\usepackage{latexsym}
\usepackage{stfloats}
\usepackage[T1]{fontenc}      % good fonts     (pdfLaTeX)
\usepackage[utf8]{inputenc}   % accept UTF-8   (pdfLaTeX)
\usepackage{graphicx}         % \includegraphics
\usepackage{caption}
\usepackage{soul}
\usepackage{amsmath,amsfonts,amssymb}
\usepackage{subcaption}
\usepackage{hyperref}
\usepackage{url}
\usepackage{wrapfig}
\usepackage{booktabs}
\usepackage[capitalize]{cleveref}
\usepackage{xspace}
\usepackage{multirow}
\usepackage[normalem]{ulem}
\usepackage{cuted}
\usepackage[compact]{titlesec}
\usepackage{makecell}
\usepackage{siunitx}
\usepackage{appendix}

\AtBeginEnvironment{appendices}{\crefalias{section}{appendix}
\crefalias{sections}{appendices}}

\usepackage{lineno}
\usepackage{xcolor}
\definecolor{Goldenrod}{RGB}{218,165,32}
\definecolor{pelicanpink}{RGB}{236,28,178}
\definecolor{poolblue}{HTML}{4a7fe3}
\definecolor{tablepurple}{HTML}{ba45dd}
\definecolor{myred}{HTML}{D72638}
\definecolor{myblue}{HTML}{1971c2}
\definecolor{mygreen}{HTML}{4F7942}
\definecolor{greenpool}{HTML}{79bd6d}
\definecolor{tokenpurple}{RGB}{94, 45, 255}
\definecolor{boldblue}{RGB}{37, 71, 139}
\definecolor{coralpink}{RGB}{249, 119, 114}
\definecolor{grassgreen}{RGB}{124, 194, 73}
\definecolor{lightpinkbg}{HTML}{F8DADA}
\definecolor{darkpinktext}{HTML}{E6B3B3}
\definecolor{brightpink}{HTML}{B0006A}
\definecolor{darkblue}{rgb}{0, 0, 0.5}
\hypersetup{colorlinks=true, citecolor=darkblue, linkcolor=darkblue, urlcolor=darkblue}

\newcommand{\resolved}[1]{}

\usepackage{microtype}

\usepackage{inconsolata}

\title{Follow the Flow: On Information Flow Across Textual Tokens in Text-to-Image Models}

\author{
Guy Kaplan\textsuperscript{1}\thanks{Equal contribution.},
Michael Toker\textsuperscript{2}\footnotemark[1],
Yuval Reif\textsuperscript{1},
Yonatan Belinkov\textsuperscript{2,3},
Roy Schwartz\textsuperscript{1}\\[1ex]
\textsuperscript{1}Hebrew University of Jerusalem\\
\texttt{\{guy.kaplan3,yuval.reif,roy.schwartz1\}@mail.huji.ac.il}\\[1ex]
\textsuperscript{2}Technion -- Israel Institute of Technology\\
\texttt{tok@campus.technion.ac.il}, \texttt{belinkov@technion.ac.il}\\[1ex]
\textsuperscript{3}Kempner Institute, Harvard University
}

\begin{document}

\maketitle
\begin{abstract}
Text-to-image generation models suffer from alignment problems, where generated images fail to accurately capture the objects and relations in the text prompt. Prior work has focused on improving alignment by refining the diffusion process, ignoring the role of the text encoder, which guides the diffusion. In this work, we investigate how semantic information is distributed across token representations in text-to-image prompts, analyzing it at two levels:{ }{ }{ }{ }{ }(1) \emph{in-item representation}—whether individual tokens represent their lexical item (i.e., a word or expression conveying a single concept), and (2) \emph{cross-item interaction}---whether information flows between tokens of different lexical items. We use patching techniques to uncover encoding patterns, and find that information is usually concentrated in only one or two of the item's tokens; for example, in the item ``San Francisco's Golden Gate Bridge'', the token ``Gate'' sufficiently captures the entire expression while the other tokens could effectively be discarded. Lexical items also tend to remain isolated; for instance, in the prompt ``a green dog'', the token ``dog'' encodes no visual information about ``green''. However, in some cases, items do influence each other's representation, often leading to misinterpretations---e.g., in the prompt ``a pool by a table'', the token ``pool'' represents a ``pool table'' after contextualization.
Our findings highlight the critical role of token-level encoding in image generation, and demonstrate that simple interventions at the encoding stage can substantially improve alignment and generation quality.\footnote{Project repository: \href{https://github.com/tokeron/lens}{\texttt{https://github.com/tokeron/lens}}}
\end{abstract}

\begin{figure*}
  \centering
  \includegraphics[width=.78\textwidth]{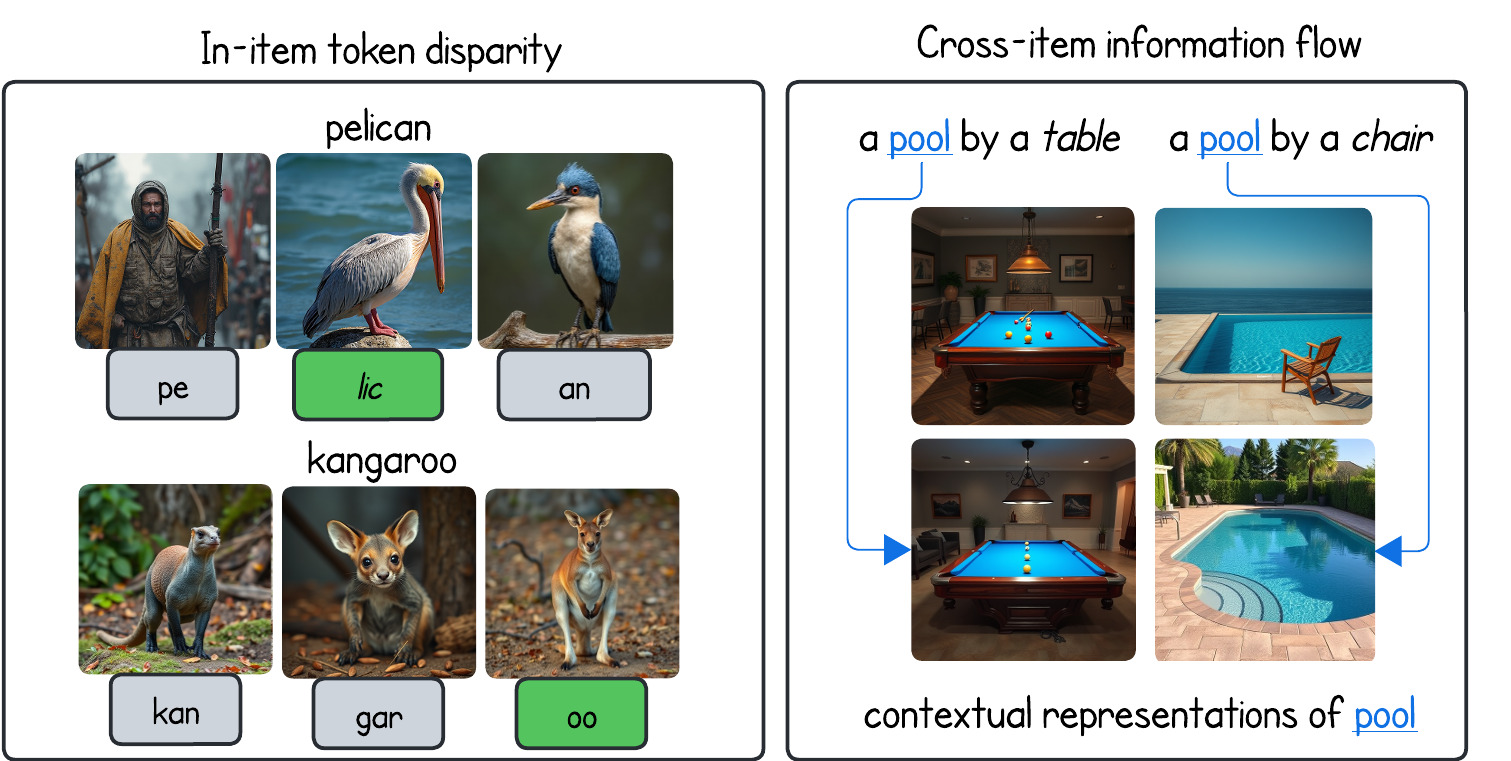}
  \caption{\textbf{Our main findings}. %
    \textbf{Left:} When generating an image based on a single input token, we find that information within a lexical item is unevenly distributed across its
    tokens' contextualized representations. In this example, one token carries the meaning of the entire item (e.g., \textcolor{greenpool}{\textbf{lic}} represents a pelican, while \textcolor{gray}{\textbf{pe}} and \textcolor{gray}{\textbf{an}} do not).
    \textbf{Right:} Contextual items may distort a token’s encoding, leading to misaligned interpretations in the generated image. 
Top: generations from the full prompts ``a \textcolor{poolblue}{\textbf{pool}} by a \textit{table}'' (misaligned) and ``a \textcolor{poolblue}{\textbf{pool}} by a \textit{chair}'' (aligned).
Bottom: generations from the contextualized token \textcolor{poolblue}{\textbf{pool}} alone. 
With \textit{table}, \textcolor{poolblue}{\textbf{pool}} encodes a pool table; with \textit{chair}, it retains the intended swimming pool meaning.
}\label{fig:main_findings}
\end{figure*}

\section{Introduction}
\label{sec:intro}
% paragraph 1: The role of the text 
Text-to-image (T2I) models typically consist of two main components: a text encoder and a diffusion model~\citep{ho2020denoising,song2019generative}. The former processes the user’s prompt, transforming it into a representation that guides the latter in generating the image. Though widely used, T2I models often exhibit prompt-image misalignment, where generated images fail to capture key concepts from the user’s prompt~\citep{chefer2023attend,rassin-etal-2022-dalle,Huang2023T2ICompBenchAE}. Prior work has attempted to address these issues by modifying the diffusion stage, and particularly the cross-attention mechanism~\citep{rassin2023linguistic,chefer2023attend,dahary2024yourself}, under the implicit assumption that each textual token reliably encodes the item it is intended to convey. This raises two fundamental questions regarding \textit{textual encoding}: (1) is the meaning of a lexical item evenly distributed across its tokens, or concentrated in just one or two? and (2) does each token exclusively encode its lexical item, or can it also absorb information from surrounding items?

In this work, we examine these questions by studying how information is distributed across tokens after the textual encoding stage. We focus on \emph{lexical items}---words or phrases that convey a single meaning, such as ``pelican'' or ``golden gate bridge''. We trace how item information is distributed both \emph{within} the tokens of a single item~(\emph{in-item}), and \emph{across} tokens of different items~(\emph{cross-item}), using the same causal framework for both analyses.\footnote{See \cref{fig:main_findings} for examples of the different cases.} To do so, we use a causal intervention framework~\citep{toker2025paddingtonemechanisticanalysis} that assesses the information encoded in each contextual token representation at the encoder's output~(\secref{method}).\footnote{Throughout this work, we study token representations at the output of the text encoder. For brevity, we sometimes omit the word ``representation''.} We then evaluate this framework on prompts drawn from widely used T2I benchmarks~(\secref{experimental_setup}).

For \intra~(\secref{Intra-concepts}), we find that a lexical item’s meaning is typically concentrated in one or two \textit{representative tokens} (tokens that alone suffice to convey the full item, e.g., ``lic'' for ``pelican'' in \cref{fig:main_findings}). Surprisingly, ablating the non-representative tokens not only does not hurt performance, but actually \emph{improves alignment} by 21\% relatively. We further show that representative tokens can be identified efficiently without image generation, opening the door to pruning non-representative tokens directly within T2I pipelines.

For~\inter~(\secref{inter_item}), we apply the same framework to ask whether information flows between different lexical items in a prompt. We observe cross-item flow in 11\% of cases. Interestingly, this flow does not always follow syntactic structure, and can result in incorrect item resolution---especially with polysemous words. For instance, in the prompt ``a \textit{pool} by a \textit{table}'', \textit{pool} can wrongly suggest \textit{table} refers to a billiard table, an instance of \textit{semantic leakage}~\citep{rassin-etal-2022-dalle}.

Overall, our analysis shows that token-level encoding is typically concentrated, largely item-isolated, yet vulnerable to \textit{semantic leakage} in cases such as polysemous words, which can lead to misinterpretations. Across our experiments, we also demonstrate simple interventions for improving alignment, highlighting the importance of further investigating the text encoder’s role in T2I.

\section{Methodology}
\label{sec:method}
We introduce a method for causal intervention by generating images from arbitrary subsets of token representations while masking the others.\footnote{While tokens interact throughout encoding, their final representations can be examined in isolation. This allows us to visualize how each token---following the encoding step---contributes to the diffusion process.}
We use it to analyze two phenomena: (1) how information is distributed across tokens within a single lexical item (\secref{Intra-concepts}), and (2) how different lexical items influence one another (\secref{inter_item}). 

\begin{figure*}[t]
  \centering
  \includegraphics[width=\textwidth]{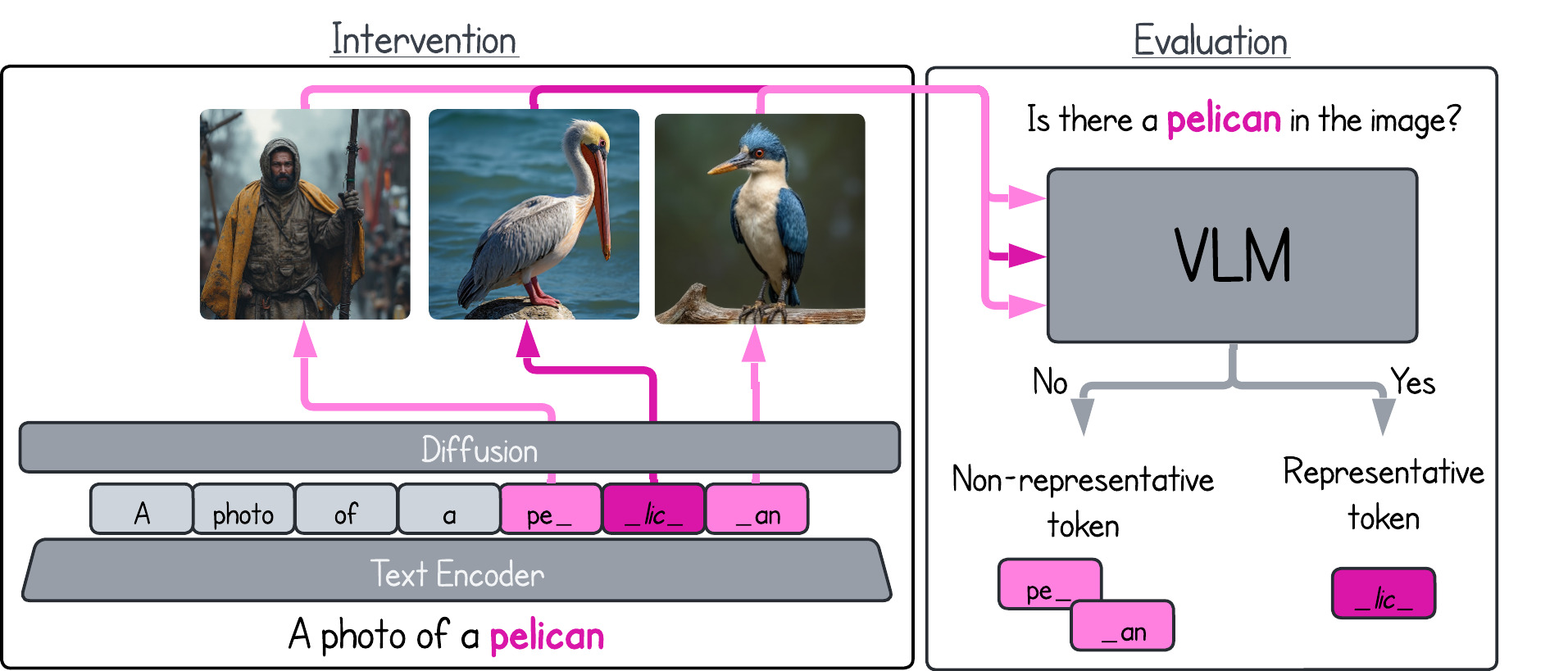}
  \caption{\textbf{Evaluating in-item information flow.}
           Our proposed framework interprets the information flow
           within a lexical item. We generate images from each token
           comprising the lexical item (left) and analyze them with a
           VLM~(right). In this example, only the token
           \textcolor{brightpink}{\textbf{lic}} represents the concept
           ``pelican'', whereas \textcolor{pelicanpink}{\textbf{pe}} and
           \textcolor{pelicanpink}{\textbf{an}} do not.}
  \label{fig:framework_in}
\end{figure*}

Our method builds on the framework proposed by~\citet{toker2025paddingtonemechanisticanalysis}.
Given a prompt with \(N\) tokens \(t_1, t_2, \ldots, t_N \), our goal is to isolate and interpret the information encoded by a subset of these tokens. Let \( S \subset \{1, \ldots, N\} \) be the indeces of a given subset, where \( 0 < |S| < N \).
We begin by encoding the full prompt using the text encoder \( E \), yielding the final hidden states \( h_1, \ldots, h_N \). Separately, we encode a sequence consisting entirely of pad tokens to obtain pad embeddings \( p_1, \ldots, p_N \). We then construct a \textit{patched prompt} by replacing all hidden states outside \( S \) with the corresponding pad embeddings:
\[
\tilde{t}_i = 
\begin{cases}
h_i &  i \in S, \\
p_i & \text{otherwise}
\end{cases}
\quad \text{for } i = 1, \ldots, N.
\]
The patched sequence \( \tilde{t}_1, \ldots, \tilde{t}_N \) is then used to guide the diffusion model. Generating an image from this patched representation allows us to isolate the individual contributions of the selected tokens as interpreted by the diffusion model.
To evaluate the information in the generated images, we use a vision-language model (VLM). See~\cref{fig:framework_in} for an illustration of our method. Below we describe our experimental setup, followed by the two phenomena we analyze using our method.

\section{Experimental Setup}
\label{sec:experimental_setup}
\paragraph{Models.}

We experiment with four text-to-image models: FLUX-schnell, FLUX-dev~\citep{flux2024}, SDXL-Turbo~\citep{sauer2023adversarialdiffusiondistillation}, and SANA~\citep{xie2024sana}. 
The FLUX models employ T5-XXL~\citep{raffel2023exploringlimitstransferlearning} as their main text encoder, SDXL-Turbo uses CLIP~\citep{radford2021learning}, and SANA uses Gemma~\citep{team2024gemma}. 
While our core insights and results are consistent across all models, certain differences arise due to encoder-specific artifacts. 
In particular, the unidirectional information flow in Gemma and CLIP, as well as the presence of a dominant [CLS] token in CLIP, lead to slightly different behaviors compared to the bidirectional T5 encoder. 
For clarity and simplicity, we focus on results from FLUX-schnell in the main paper, and discuss these cross-model differences in detail in \secref{discussion}.

\paragraph{Data.}
\label{sec:data}
We use a subset of 1{,}053 prompts from the DrawBench~\citep{saharia2022photorealistic} and PartiPrompts~\citep{yu2022scaling} datasets, filtered to include 4–20 words prompts and exclude cases with added complexity unrelated to our focus (e.g., misspellings, written text, rare words). For each prompt, we generate five images using different random seeds. 
To extract the lexical items, we prompt \mbox{GPT-4o}~\citep{openai2024gpt4technicalreport}, resulting in 4{,}864 unique items across prompts. See~\appref{app:models:lexical_items_cls} for further technical details. We then use spaCy~\citep{spacy} to determine the part-of-speech of each lexical item, and retain only nouns, proper nouns, and adjectives, as these are typically concrete and can be identified in their visual representation. We end up with 3{,}891 unique lexical items and use them to evaluate both~\intra~and~\inter.

\paragraph{Evaluation.}
To evaluate generated images, we employ Qwen2-VL-72B-Instruct~\citep{Qwen2VL} 
using binary (yes/no) questions to assess prompt-image alignment and item 
presence~(see~\appref{app:vlms} for details).While recent work shows that VLMs still struggle significantly with complex visual abductive reasoning \citep{venturanl}, we find that for basic perceptual tasks like item presence, model predictions align well with human judgment. To validate its reliability, 
three human annotators evaluate 100 randomly sampled cases, evenly split 
between~\intra~(\secref{Intra-concepts}) and~\inter~(\secref{inter_item}) 
settings, yielding substantial agreement (Cohen's Kappa of 0.868 and 0.764, 
respectively). Model predictions align well with human judgment, with 
accuracy/F1 of 0.927/0.933 for~\intra~and 0.810/0.740 for~\inter.

\section{In-Item Representation}
\label{sec:Intra-concepts}
In this section, we analyze information flow within lexical items at the token level. We first show semantic information is unevenly distributed across tokens, with typically one or two tokens representing the item~(\secref{how_is_info_distributed}). We then show non-representative tokens are largely redundant and may even harm generation~(\secref{redundant_tokens}).

\subsection{Information Distribution Across Tokens}
\label{sec:how_is_info_distributed}
We begin by exploring how information is distributed across the tokens within a lexical item. Is it spread evenly across tokens, or rather concentrated in specific ones? This question is of interest as many T2I applications treat all tokens in a given prompt equally~\citep{chefer2023attend, rassin2023linguistic, dahary2024yourself}. Uncovering an asymmetrical distribution of information could improve the effectiveness of such applications by focusing on the most informative tokens.

\begin{figure}[t]
  \centering
  \includegraphics[width=\linewidth]{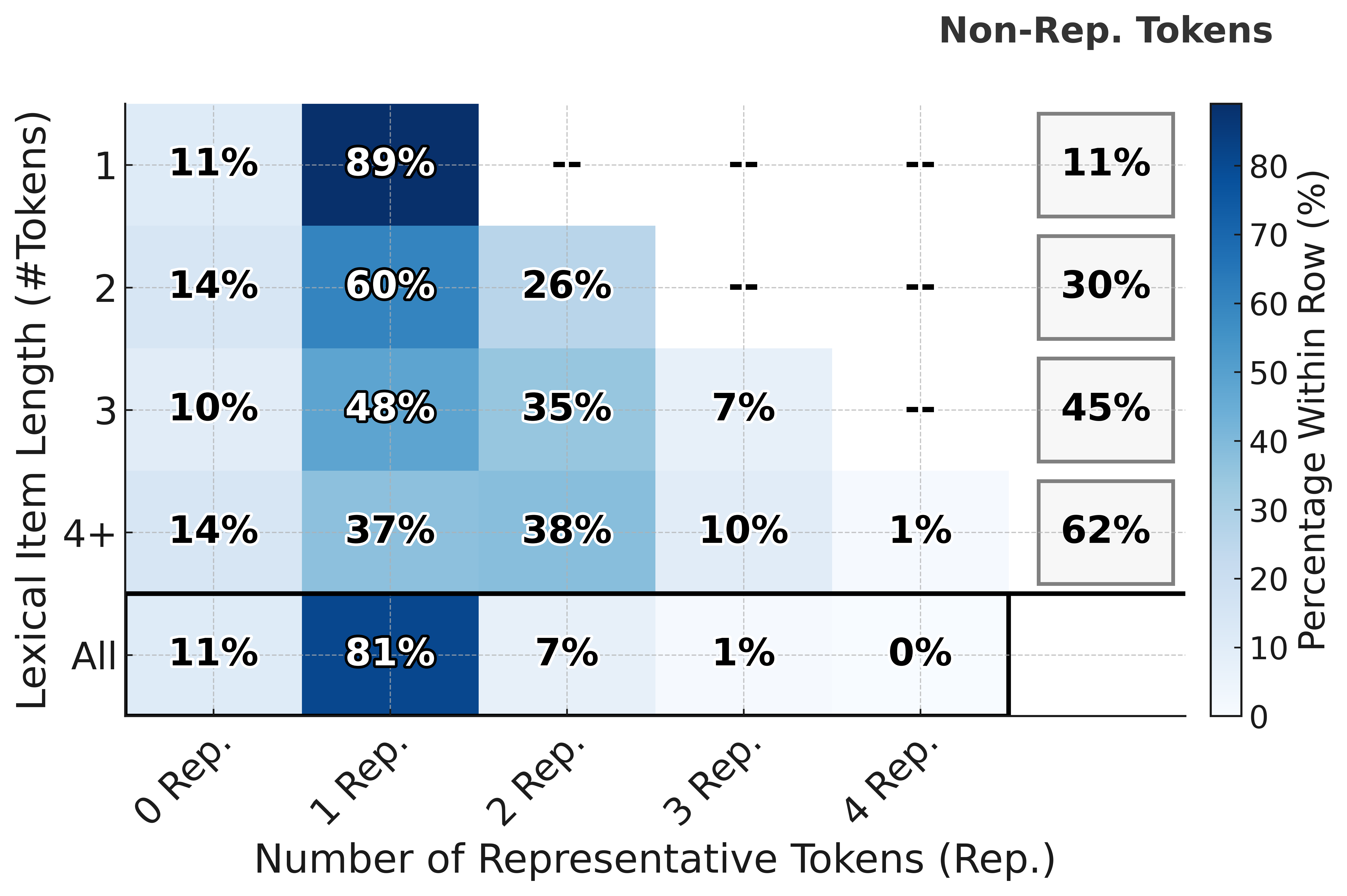}
  \caption{Distribution of representative tokens per item length. ``Rep.'' denotes the number of representative tokens; the bottom row aggregates over all lexical items. In most cases, one or two tokens are sufficient to represent the entire item. As item length increases, the number of non-representative tokens grows accordingly.
  }
  \label{fig:rep-token-dist}
  % \vspace{-0.5cm}
\end{figure}

Given a prompt and a lexical item, we feed the prompt to the text encoder, obtaining contextualized token representations. We then identify the tokens comprising the item and apply our intervention method~(\secref{method}) to generate an image conditioned on each token's representation. Our goal is to study how much of the item’s meaning is retained in that token's contextual representation, as probed from the final encoder layer. Finally, we use Qwen2-VL to assess whether the image represents the overall lexical item it is part of.

We define a \textit{representative token} as one whose isolated representation results in an image containing its corresponding lexical item. Tokens that do not meet this criterion are considered \textit{non-representative}. We repeat this analysis for each lexical item and for each prompt in our dataset. For example, given the prompt ``a \textit{spaceship} that looks like the \textit{Sydney Opera House},'' we conduct the experiment for each of ``spaceship's'' tokens (`space', `ship') and ``Sydney Opera House'' tokens (`Sydney', `Opera' and `House').
Our results~(\cref{fig:rep-token-dist}, \textsl{bottom row}) show that in 89\% of the cases, there is at least one \textit{representative token}. Interestingly, in cases where no such tokens exist, the item is usually absent from the full-prompt image as well, partly reflecting gaps in the encoding process~(see \appref{app:neglect} for detailed analysis and examples).

\begin{table*}[t]
\centering
\scriptsize
\resizebox{\textwidth}{!}{%
\begin{tabular}{
    l
    c
    S[table-format=2.2]
    S[table-format=2.2]
    S[table-format=3.2]
    S[table-format=1.2]
    S[table-format=2.2]
    S[table-format=+1.2, retain-explicit-plus]
}
\toprule
& & \multicolumn{2}{c}{\textbf{Acc.}} & \multicolumn{3}{c}{\textbf{Relative \%}} & \\
\cmidrule(lr){3-4} \cmidrule(lr){5-7}
\textbf{Tokens} & & & & & & & \\
\textbf{Removed} &
{\textbf{\# Prompts}} &
{\textbf{Before}} &
{\textbf{After}} &
{\textbf{Unaffected}} &
{\textbf{Degraded}} &
{\textbf{Improved}} &
{\textbf{Accuracy $\Delta$ (pp / rel.)}} \\
\midrule
1   & 144 & 81.25 & 83.33 &  98.29 & 0.85 & 14.83 & +2.08\ (11\%) \\
2   &  98 & 82.65 & 88.78 & 100.00 & 0.00 & 35.27 & +6.13\ (35\%) \\
3   &  45 & 93.33 & 93.33 &  97.62 & 2.38 & 33.28 & +0.00\ (0\%) \\
4   &  24 & 87.50 & 91.67 & 100.00 & 0.00 & 33.36 & +4.17\ (33\%) \\
5+  &  28 & 78.57 & 85.71 & 100.00 & 0.00 & 33.32 & +7.14\ (33\%) \\
\midrule
\textbf{Overall} & 339 & 83.48 & 87.02 & 98.90 & 0.71 & 25.00 & +3.54\ (21\%) \\
\bottomrule
\end{tabular}}
\caption{Effect of removing non-representative tokens on prompt-level accuracy, reporting accuracy before and after removal, and the percentage of prompts that remained successful~(Unaffected), became failures~(Degraded), or were corrected~(Improved). Removing non-representative tokens rarely harms generation and often improves it, yielding an overall 21\% relative reduction in generation errors overall.}
\label{tab:token-removal}
\end{table*}

We next focus on instances where at least one token represents the lexical item, and examine the number of \textit{representative} and \textit{non-representative tokens} across items of different lengths. Our results~(\cref{fig:rep-token-dist}, \textsl{top rows}) show that typically one or two tokens represent the concept, while the remaining tokens are \textit{non-representative}. Further, as the token length of the lexical item becomes longer, the number of \textit{non-representative} tokens increases~(\cref{fig:rep-token-dist}, \textsl{rightmost column}). 
Inspecting all lexical items in our data composed of two or more tokens, these \textit{non-representative tokens} account for 52\% of their tokens. We next examine the effect of removing \textit{non-representative tokens} altogether.

\begin{figure*}[t]
  \centering
  \includegraphics[width=\textwidth]{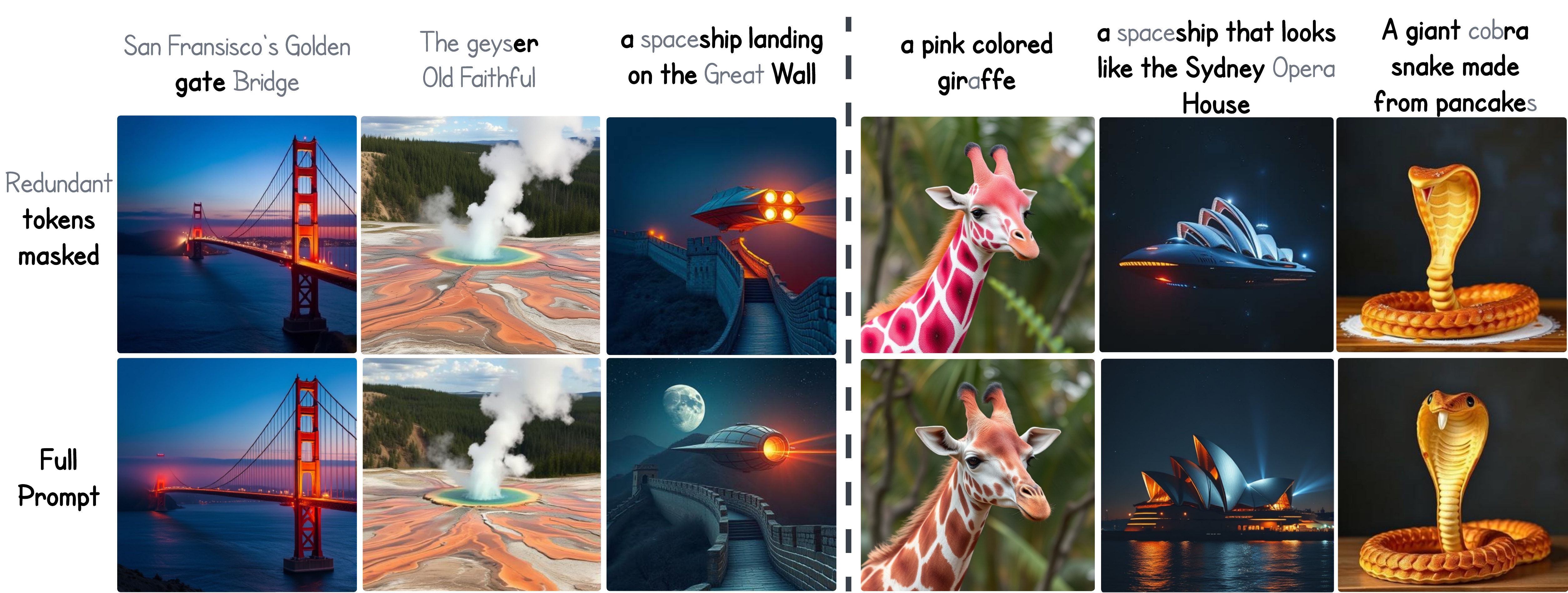}
  \caption{Examples illustrating the effect of removing \textit{non-representative tokens}. \textbf{Top row}: Images generated after removing \textit{non-representative tokens} (Representative tokens are shown in \textbf{bold}; non-representative tokens are in \textcolor{gray}{gray}).  
\textbf{Bottom row}: Images generated from the full prompt.  
\textbf{Left}: In most cases, removal results in no noticeable effect on the generation.
\textbf{Right}: In some cases, removal improves alignment with the prompt.
}
  \label{fig:redundancy_removal_examples}
\end{figure*}

\subsection{Textual Tokens During Diffusion: Are They All Necessary?}
\label{sec:redundant_tokens}
We now examine whether \textit{non-representative} tokens have a prominent effect on image generation. To answer this question, we apply our intervention method to each prompt, this time generating an image after masking all the \textit{non-representative tokens}, while keeping \emph{all representative tokens}. We use Qwen2-VL to measure whether this image aligns with the prompt, and compare it to an image generated from the full prompt without intervention.

\paragraph{\textit{Non-representative tokens} are redundant.} Our results (\cref{tab:token-removal}) show that removing \textit{non-representative tokens} generally preserves quality~(see \cref{fig:redundancy_removal_examples}, left-hand side for illustration). When the original generation is aligned with the prompt, the generated image after non-representative token removal remains aligned in 98\% of cases, suggesting these tokens are largely redundant. Surprisingly, in cases where the original image fails to align with the prompt, we observe a 21\% relative improvement in from 16.5\% to 13\% failure rate) alignment after removing non-representative tokens
(see \cref{fig:redundancy_removal_examples}, right-hand side). We attribute this improvement to the model relying exclusively on the remaining representative tokens, which encode the correct semantics of the item.\footnote{We measure improvement rates with Qwen2-VL before and after masking non-representative tokens (\appref{app:vlms}).}

\subsection{Efficient Redundant Token Identification}
\label{sec:practical}
While identifying and removing redundant tokens before the diffusion process improves image-text alignment, this identification process is computationally expensive, as it requires generating an image from each subtoken of a lexical item.

To make this process efficient, we introduce a lightweight classifier that predicts token redundancy directly from text encoder representations, \textit{without generating any images}. 
This classifier operates immediately after textual encoding, allowing redundant tokens to be masked before the prompt is passed to the diffusion model. 
It enables fast, on-the-fly filtering and straightforward integration into existing T2I pipelines.
Using the same dataset of token-level annotations described (after the classification using the original patching technique), we train a lightweight single linear layer classifier to predict redundancy from token embeddings~(see~\appref{appendix:classifiers} for details). As shown in \cref{tab:knn-results}, the classifier achieves \textbf{90\% precision} and 83\% accuracy, ensuring that representative tokens are rarely misclassified and inadvertently removed.

\begin{table}[h]
\centering
\resizebox{\linewidth}{!}{%
\begin{tabular}{lcccc}
\toprule
\textbf{Metric} & \textbf{Accuracy} & \textbf{Precision} & \textbf{Recall} & \textbf{F1-score} \\
\midrule
\textbf{Score}  & 0.83 & 0.90 & 0.86 & 0.88              \\
\bottomrule
\end{tabular}%
}
\caption{Predicting token redundancy by encoded representations alone.}
\label{tab:knn-results}
\end{table}

By filtering out redundant tokens before generation, the diffusion model receives cleaner, more focused inputs, improving prompt-image alignment and reducing noise. 
Using only representative tokens yields a \textit{21\% reduction in generation failures}~(Table~\ref{tab:token-removal}, Figure~\ref{fig:redundancy_removal_examples}).
This transition moves our approach beyond analysis, establishing it as a practical component that can be applied to efficiently enhance the overall alignment of T2I pipelines.

\begin{table}[t]
\centering
\small
\begin{tabular}{lrr}
\toprule
\textbf{Category} & \textbf{Count} & \textbf{Percentage} \\
\midrule
\# pairs                       & 15{,}950 & 100.00\% \\
No Information Flow               & 14{,}251 & 89.35\%  \\
Information Flow             & 1{,}699  & 10.65\%  \\
\quad -- Source before Reference  & \phantom{0}835   & 49.15\%  \\
\quad -- Reference before Source  & \phantom{0}823   & 48.44\%  \\
\bottomrule
\end{tabular}
\caption{Distribution of information flow between lexical item pairs. Subcategories under ``Information Flow'' indicate the order of source and reference.}
\label{tab:info-flow-stats}
\end{table}

\section{Cross-Item Interactions}
\label{sec:inter_item}

In~\secref{Intra-concepts}, we have demonstrated that a single token can effectively encapsulate the semantics of an entire lexical item. This concentration makes the representative token especially critical — if cross-item information flow contaminates it, the entire concept may be misrepresented. Applying the same causal intervention framework at a broader scope, we now ask: what defines the segmentation boundaries of a lexical item? Can its tokens also encode information about adjacent items, or is their meaning primarily confined within the item itself?

To this end, we isolate each lexical item in a prompt and assess whether it encodes information about \emph{other} items in the prompt. For each item, we generate images from its contextualized representation (encoded within the full prompt), and its uncontextualized representation (using the item's text by itself as a prompt). For example, for the prompt ``tree \textit{reflection} in the lake'', we generate two images: one using the single word \textit{reflection}, without any context, and another where the same item is contextualized as part of the full prompt. We then use Qwen2-VL to evaluate whether any other item from the prompt---such as \textit{tree} or \textit{lake}--- appears in the contextualized image, but not in the uncontextualized one, which would indicate flow introduced by the surrounding context.
See~\cref{app:intervention_inter} for implementation details.

\begin{figure}[t]
  \centering
  \includegraphics[width=\linewidth]{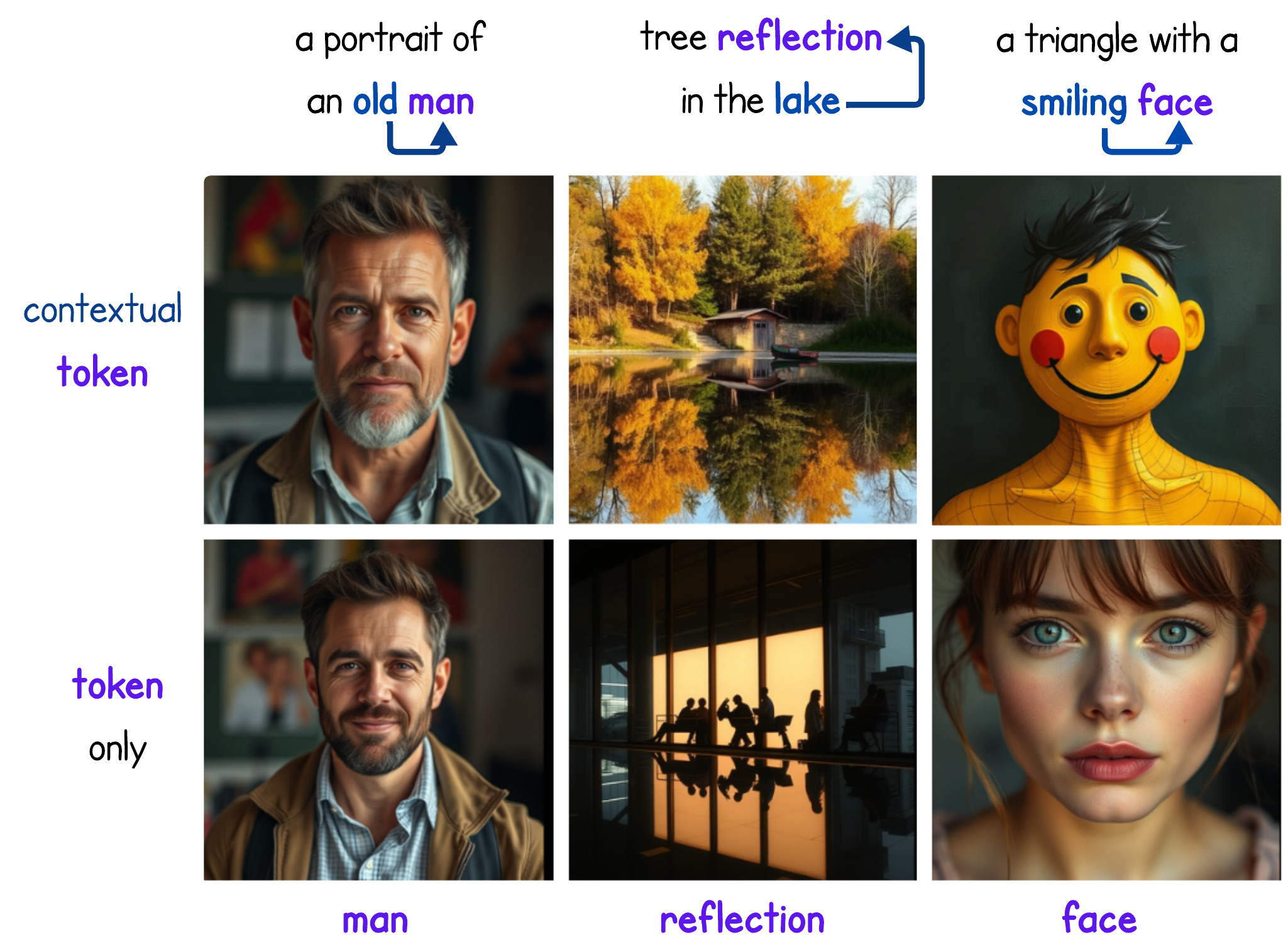}
  \caption{\textbf{Examples of information flow between items.}
  Top: Images generated from a \textbf{\textcolor{tokenpurple}{lexical item}} encoded alongside \textbf{\textcolor{boldblue}{another item}} that alters its representation. Bottom: Images generated from the uncontextualized representation of the same lexical item. 
  }
  \label{fig:inter_token_flow}
\end{figure}

Our results~(\cref{tab:info-flow-stats}) show that in general, lexical items do not encode information from other items in the prompt: 89\% of the tokens do not lead to images containing other items. Interestingly, in the remaining 11\%, information flow can also emerge between items with no direct syntactic relation in the prompt---e.g., information flowing from \textit{lake} into the representation of \textit{reflection}.\footnote{See \cref{fig:inter_token_flow} for qualitative examples.}

\begin{figure*}[t]
  \centering
  \includegraphics[width=\linewidth]{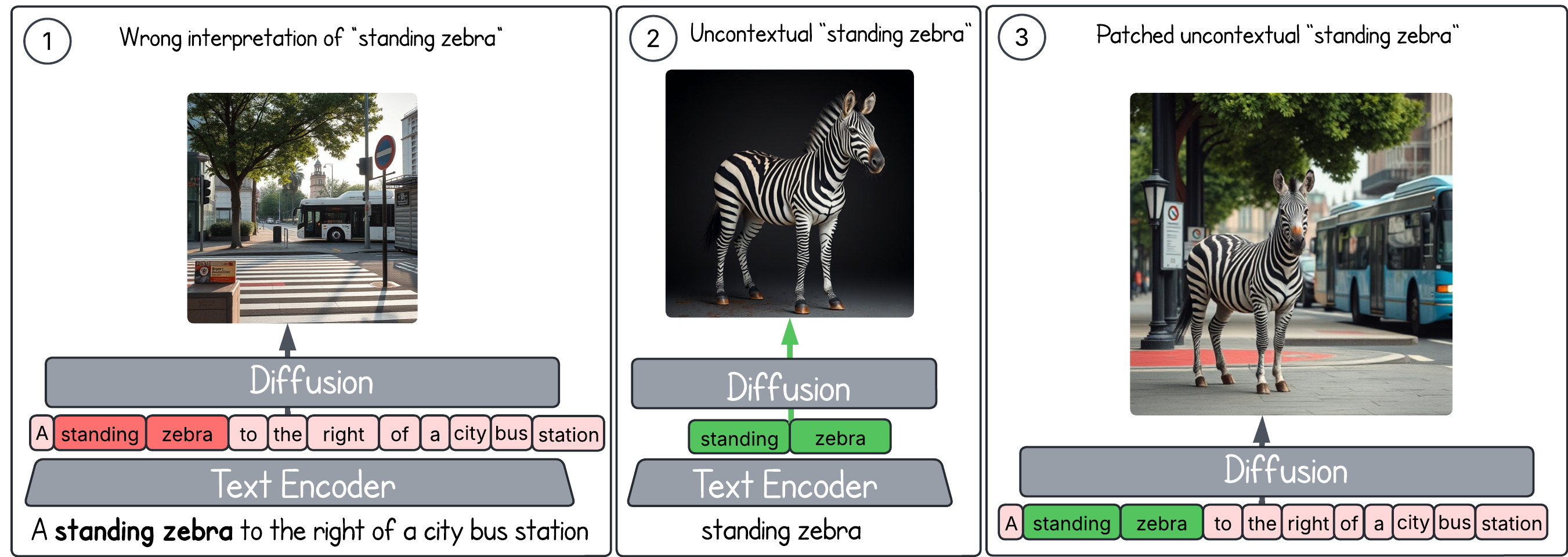}
    \caption{\textbf{Verifying\ and mitigating textual semantic leakage by replacing the contextually leaked concept representation.} (1) Regular generation produces an image showing a crosswalk to the right of a bus station.{ }{ }{ }{ }{ }{ }{ }{ }{ }{ }(2) Generation from the prompt \textcolor{grassgreen}{``standing zebra''}, without any context, results in the expected interpretation of the zebra as an animal. (3) Mitigating the leakage by generation using the original prompt, but with the leaked concept \textcolor{coralpink}{``standing zebra''} replaced by its \textcolor{grassgreen}{uncontextualized representation}.}
  \label{fig:application_clean}
\end{figure*}

\subsection{Can Information Flow Lead to Misinterpretation of Items?}
\label{par:cleaning application}

Our analysis reveals a recurring pattern of semantic influence between items, where the interpretation of one item can be affected by another that is syntactically distant.
While such influence is crucial, as language is inherently contextual, sometimes this influence becomes misleading or distorting.
For example, in ``a standing \textit{zebra} to the right of a city \textit{bus station},'' the association between \textit{bus station} and \textit{zebra} causes the model to depict a \textit{zebra crossing} instead of the animal (see \cref{fig:application_clean}). We refer to this \textit{clearly unambiguous}, unintended and misleading form of cross-item influence as \textit{semantic leakage}~\citep{rassin-etal-2022-dalle, gonen2024does}.  We note that this definition is behaviorally defined---a leakage only occurs if the model is wrong in a clearly unambiguous case.

We note that such distorting forms of information flow typically arise in cases involving polysemous words. For example, the item \textit{bats} is interpreted as \textit{baseball bats} in the prompt ``bats fly around a baseball stadium'', even though the intended meaning is obviously the flying animal. We focus on cases in which the context provides a \textit{definitive cue} for resolving ambiguity of the polysemous word, but the model still fails to resolve the item correctly.
We hypothesize that these cases are caused by \textit{semantic leakage} during the textual encoding, where unintended associations override contextually appropriate meaning. 

To test this hypothesis, we use the dataset from \citet{rassin-etal-2022-dalle}, which includes prompts known to induce misalignment due to implicit lexical associations (e.g., \textit{bat} and \textit{baseball stadium}). Unlike standard T2I benchmarks such as \citet{huang2023t2icompbench}, which focus on spatial or visual challenges, these prompts highlight failures rooted in the encoding process.  Since the original set includes only 30 relevant examples, we expand it with GPT-4o to generate additional candidates.
We then keep only prompts whose intended interpretation of each polysemous word was judged by human annotators to be \textit{fully unambiguous}, and compare that intended meaning to the one expressed in the generated image. Prompts where the model’s output contradicts the intended sense produce a final curated set of 110 examples.\footnote{See Appendices \ref{app:intended_pipeline} and \ref{app:leakage_prompts} for more details.}

On those prompts, we compare two generations: (1) from the full prompt, (2) from a ``cleaned'' version containing only the item and its modifiers. 
For example, given the prompt ``\textit{bats} fly around a baseball stadium'', we generate images of \textit{bats} once encoded in the phrase ``\textit{bats} fly'' and once encoded in the full context. 
Each image is then evaluated for whether it depicts the intended interpretation of the item---\textit{an animal bat}~(see \cref{fig:application_clean} for details). 
In 93\% of cases, the item is incorrectly resolved in the contextualized item level, indicating that the error originates at the encoding stage, whereas the cleaned version preserves the intended meaning.

\subsection{Mitigation of Semantic Leakage}
\label{sec:mitigation}
Having established that misalignment may arise from \textit{semantic leakage} during encoding, we now turn to a method for mitigating it. We apply a patching technique (\cref{fig:application_clean}, step 3), in which the wrongly resolved items are replaced with their ``cleaned'' representation. Importantly, when constructing this cleaned representation, we encode the item together with its local modifiers (e.g., \textit{standing zebra} rather than \textit{zebra} alone), preserving fine-grained context relevant for interpretation while correcting the cross-item leakage. Our method resembles \citet{feng2023trainingfreestructureddiffusionguidance}, but differs in that we replace the target tokens completely, and not just the value projection. We use RAG-Diffusion~\citep{tan2024ragdiffusionfaithfulclothgeneration} as a baseline, which improves alignment by creating bounding boxes for each item and restricting diffusion attention to those areas.

As shown in Table~\ref{tab:semantic_leakage}, our patching technique consistently achieves the strongest reduction, lowering misinterpretations to 20\% on Flux-Dev and 14\% on Flux-Schnell, substantially outperforming the baseline, demonstrating that patching provides an effective solution for mitigating \textit{semantic leakage} of this kind. While our analysis focused on prompts with clear intended meanings, similar effects arise in ambiguous (e.g., ``bats in a stadium'') or biased contexts (e.g., gender-related prompts), and persist in the closed model Nano-Banana~(\citet{geminiteam2025geminifamilyhighlycapable}; see \appref{appendix:closed_models} for more details). 
% We further discuss these cases in Appendices~\ref{app:polysemous} and~\ref{app:bias_mitigation}, showing how patching naturally extends to user-guided control and bias mitigation.

\begin{table}[htbp]
\centering
\small
\begin{tabular}{lccc}
\toprule
\textbf{Model} & \makecell{\textbf{Initial} \\ \textbf{Leakage}} 
               & \makecell{↓ \textbf{RAG-} \\ \textbf{Diffusion}} 
               & \makecell{↓ \textbf{Patching} \\ \textbf{(Ours)}} \\
\midrule
Flux-Dev      & 79\% & 39\% & \textbf{20\%} \\
Flux-Schnell  & 94\% & 43\% & \textbf{14\%} \\
\bottomrule
\end{tabular}
\caption{Mitigating semantic leakage across models. 
Values show percentage of prompts exhibiting leakage.}
\label{tab:semantic_leakage}
\end{table}

\section{Discussion}
\label{sec:discussion}

\paragraph{Beyond leakage: polysemy control and bias mitigation.}
Our mitigation approach in \secref{mitigation} replaces a misinterpreted item's 
contextualized representation with its uncontextualized, ``clean'' counterpart---effectively 
patching in the item's \textit{null-context} meaning.
This same mechanism generalizes naturally into the following two use-cases.
First, when a word is inherently polysemous and the context alone does not resolve the 
intended sense (e.g., \textit{crane} as a bird vs.\ a construction machine), a user can 
manually select the desired interpretation and patch in the corresponding contextualized 
representation, achieving the intended generation in 95\% of cases (see \appref{app:polysemous} 
for details and examples).

\begin{figure*}[t]
  \centering
  \includegraphics[width=.68\textwidth]{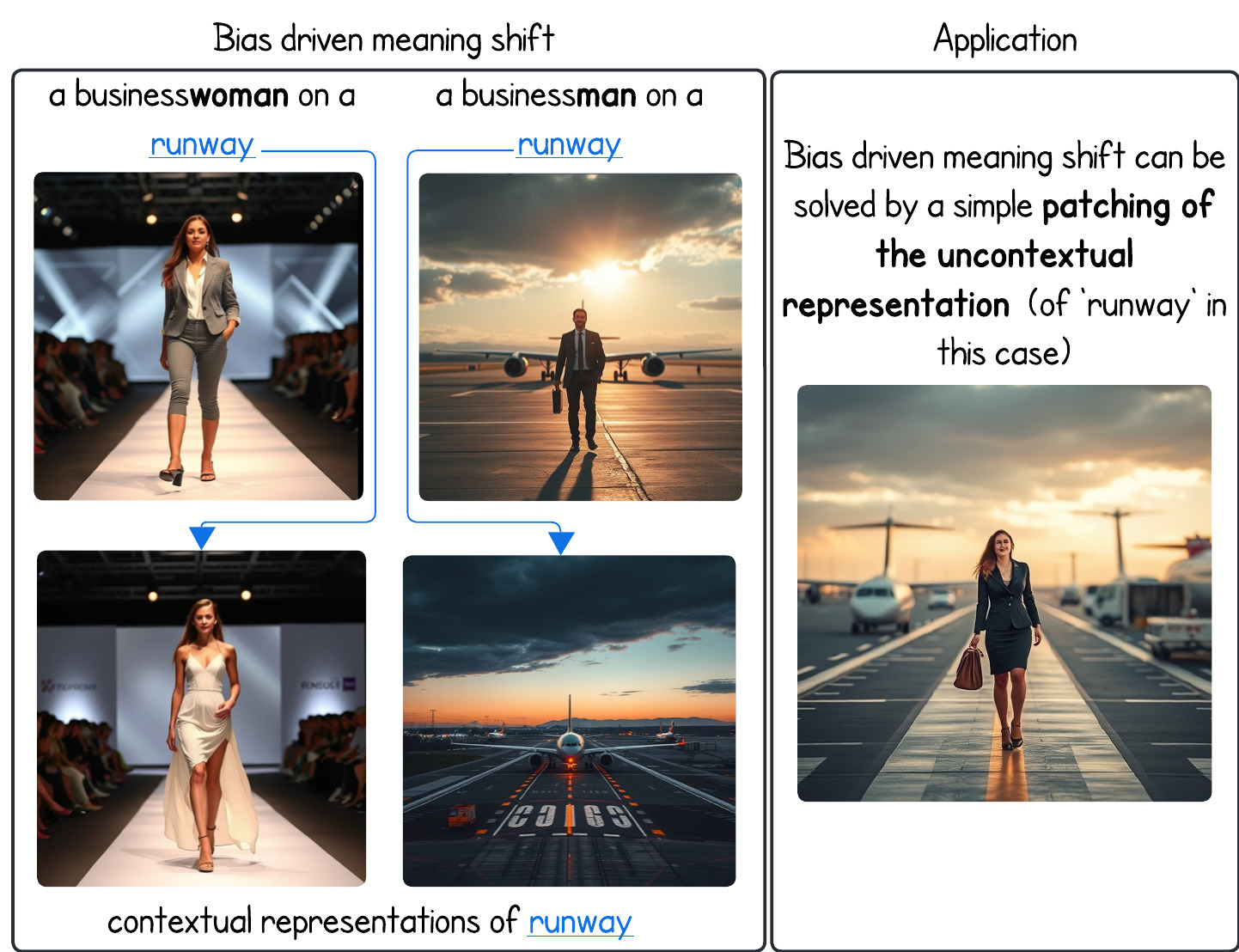}
  \caption{\textbf{Bias-driven meaning shift and mitigation.}
  \emph{Left:} The top row show generations for ``a business\textbf{woman} on a \textit{runway}'' (fashion runway) and ``a business\textbf{man} on a \textit{runway}'' (airport runway). 
  The bottom row shows generations from the contextualized token \textit{runway} alone, revealing how gender context distorts its encoding.
  \emph{Right:} Patching the \textit{runway} token with its uncontextualized representation mitigates this bias, yielding an unbiased image of a businesswoman on an airport runway.}
  \label{fig:bias_patching_runway}
\end{figure*}

Second, context can also introduce \textit{bias}. For example, in the prompts ``a 
business\textbf{woman} on a \emph{runway}'' and ``a business\textbf{man} on a 
\emph{runway}'', the gender context shifts the encoded meaning of \textit{runway} toward a 
fashion runway or an airport runway respectively~(see \cref{fig:bias_patching_runway}).
Patching \textit{runway} with its uncontextualized representation corrects this 
gender-driven drift, producing the intended airport runway regardless of the gender cue.
More broadly, this suggests that encoder-side patching is a lightweight intervention that extends naturally to user-guided control and bias mitigation.

\paragraph{Developing textually challenging benchmarks.}  
Current T2I datasets focus primarily on visual or spatial complexity~\citep{Huang2023T2ICompBenchAE,ghosh2023genevalobjectfocusedframeworkevaluating,saharia2022photorealistic, yu2022scaling}. Yet our findings show that even slight linguistic ambiguity—particularly with polysemous or compositional phrases—can cause encoding failures. This highlights the need for evaluation benchmarks that probe textual difficulty more directly, which may in turn drive improvements in encoder design.

\paragraph{Generalization across T2I architectures.}
Our analysis across three encoder types—T5 (FLUX models), Gemma (SANA), and CLIP (SDXL-Turbo)—reveals two key differences that shape semantic representation and interpretability. First, T5 is bidirectional, allowing representative tokens to appear at various positions within a lexical item, while Gemma and CLIP are unidirectional, consistently placing representative tokens at the final token. This regularity in unidirectional models simplifies practical tasks such as redundant token masking. Second, T5 and Gemma distribute meaning across multiple tokens, with over 89\% of lexical items containing at least one representative token, whereas CLIP centralizes most semantic information in its [CLS] token, leaving other tokens weak and limiting token-level interpretability. These distinctions, discussed further in Appendices~\ref{app:SDXL-Turbo_comparison} and~\ref{app:sana_comparison}, highlight how encoder architecture impacts both analysis and downstream applications.

\paragraph{What makes a token representative?}
While unidirectional encoders consistently place representativity at the final token of a lexical item, the question of \textit{which} token becomes representative in bidirectional encoders like T5 remains open. Preliminary analysis reveals only moderate or weak correlations with interpretable surface features such as edit distance to the full lexical item ($r \approx -0.4$) or pretraining co-occurrence frequency ($r \approx 0.15$), and many representative subtokens carry no obvious surface-level semantics (e.g., \textit{T} in \textit{T-shirt}). We leave a deeper investigation to future work.

\section{Related Work}
\label{sec:related}

\paragraph{Interpretability in T2I models.}
Recent work has explored how T2I models encode and align concepts, including studies of CLIP's latent space~\citep{chefer2023hidden}, 
evolution of representations across the textual encoding process~\citep{toker-etal-2024-diffusion}, alignment via attention maps~\citep{tang-etal-2023-daam}, and sparse autoencoders for intermediate representations~\citep{cunningham2023sparse,cywinski2025saeuron}. In contrast, we focus on token-level information at the encoder's final layer, which directly conditions the generation process.
Concurrent work by \citet{wang2026circuitmechanismsspatialrelation} adopts a mechanistic interpretability approach 
to study spatial relation generation in diffusion transformers, finding that T5 fuses 
information from multiple lexical items into single tokens---a phenomenon that closely 
mirrors our findings on representative tokens (\secref{Intra-concepts}) and 
cross-item information flow (\secref{inter_item}).

\paragraph{Token representation and flow in LLMs.}
Token information in LLMs is not uniformly distributed: subwords fuse into word-level meaning \citep{kaplan2025tokenswordsinnerlexicon}, and later tokens can erase earlier ones \citep{feucht2024tokenerasurefootprintimplicit}. Our findings suggest this compression extends beyond single words: in expressions like "San Francisco's Golden Gate Bridge", a single token can represent the entire phrase — consistent with the motivation behind super-word tokenization schemes such as SuperBPE (\citep{liu2025superbpespacetravellanguage}). Attention-based flow \citep{vig-belinkov-2019-analyzing,clark-etal-2019-bert} can be misleading \citep{pruthi-etal-2020-learning}, prompting alternatives like Attention Rollout \citep{abnar-zuidema-2020-quantifying} or representation decoding via Patchscopes \citep{Ghandeharioun2024PatchscopesAU} and logit lens \citep{nostalgebraist}—though these do not test whether downstream components actually use the encoded information. Our causal intervention approach addresses this gap directly, using image generation as a rich probe of what information each token representation actually carries.

\paragraph{Probing and causal methods.}
Probing methods~\citep{adi2016fine,liu-etal-2019-linguistic,zhang2018language,brunner2019identifiability} reveal information presence but are not fully reliable, as probes can exploit spurious correlations~\citep{belinkov-2022-probing}. Instead, we use causal interventions to test whether token information is used during image generation.

\paragraph{Challenges in T2I models.}
\textit{Semantic leakage}—when context distorts word meaning—occurs in both LLMs~\citep{gonen2024does} and T2I~\citep{rassin-etal-2022-dalle,dahary2024yourself}. Another common issue is \textit{neglect}, where key items are omitted~\citep{chefer2023attend,chang-etal-2024-repairing}, as well as a lack of cultural alignment, where models struggle to accurately represent cultural nuances \citep{ventura2025navigating}. While prior work focuses on solutions during the diffusion process, we show that some of 
these failures often originate in the text encoder, a finding corroborated independently 
by~\citet{zarei2025improvingcompositionalattributebinding} for the case of compositional attribute binding in CLIP.
Concurrently, \citet{ventura2025deleakerdynamicinferencetimereweighting} address semantic leakage from the diffusion side, 
complementing our encoder-side analysis.

\section{Conclusion}
We presented a causal masking framework for probing semantic content encoded in individual tokens at the text encoder's output in T2I models. We showed that (1) lexical meaning is often concentrated in one or two dominant tokens, and (2) cross-item information flow, observed in ~11\% of cases, can distort item meaning. Both insights translate into practical improvements, namely redundant token masking and representation patching, highlighting the text encoder as a key source of alignment failures and a promising target for intervention.
% These findings are causally linked: precisely because meaning concentrates in so few tokens (\secref{Intra-concepts}), cross-item leakage into those tokens (\secref{inter_item}) is especially damaging. 

\section*{Limitations}
\label{sec:limitations}

Evaluating token-level representations remains challenging. While we rely on strong vision-language models as judges and validate key findings through human evaluation, these are still approximations of true semantic alignment. Our prompt set focuses on object-centric, syntactically simple cases, which may limit generalization to prompts involving misspellings, rare words, or abstract concepts. Further work is needed to explore how information flow behaves under more linguistically complex conditions.

\section*{Acknowledgments}
\label{sec:acknowledgments}
This research was supported in part by the Israel Science Foundation (grants No.\ 2045/21 and 2942/25), NSF-BSF grant 2020793, Coefficient Giving, the European Union (ERC, Control-LM, 101165402), and by an Academic Gift by NVIDIA. Michael Toker is supported by the Azrieli Graduate Studies Fellowship .Views and opinions expressed are however those of the author(s) only and do not necessarily reflect those of the European Union or the European Research Council Executive Agency. Neither the European Union nor the granting authority can be held responsible for them.

\bibliography{custom}

\clearpage
\begin{appendices}

\appendix

\label{sec:appendix}
\section{Technical Details}
\subsection{Lexical Item Classification} 
\label{app:models:lexical_items_cls}
We define a \textit{lexical item} as either a single word or a compound expression of multiple words that, in context, conveys a unified semantic meaning. A compound expression is treated as a single item when its words form a fixed lexical unit with cohesive semantics rather than merely exhibiting a modifier–head relationship. For example, while ``broken mirror'' describes a mirror’s state, expressions like ``hot air balloon'' or ``teddy bear'' denote entities with distinct identities. Similarly, although phrases such as ``identical twins'' or ``baseball bat'' might be interpreted as separate concepts, conventional usage supports their treatment as unified entities. We employ the reasoning model \texttt{o3-mini-high} as a classifier to tag multi-word lexical items in both the target prompts and the dataset. The model returns a list of identified multi-word expressions, while the remaining untagged words are treated as individual lexical items. 

\subsection{Evaluation Visual Generations.}
\label{app:vlms}
We evaluate whether an image matches a textual description using Qwen2-VL-72B-Instruct~\citep{Qwen2VL}. The following prompt is used:
\begin{quote}
"In Yes, No and maybe. Does every image match one of those descriptions: (description string)? Answer Yes if all images match or relate to at least one description, Maybe if only some match, otherwise No."
\end{quote}
Here, the \textit{textual description} can be either a single lexical item or a complete textual prompt.

\subsection{Evaluating Generated Textual Descriptions.}
\label{app:gpt-ps}
We employ GPT-4o~\citep{openai2024gpt4technicalreport} to evaluate the textual interpretations produced by Patchscopes.
We use the following prompt:
\begin{quote}
``In Yes, No and Maybe. Does every image match the description: \{Patchscopes\_description\}~?
Answer Yes if all images match or relate to the description, Maybe if only some match, otherwise No.''
\end{quote}

\subsection{Evaluating Relations Between Items.}
\label{app:llm-judge}
We enhance our leakage validation by distinguishing between cases where two lexical items exhibiting semantic leakage are perceptually bound together---for example, ``old'' and ``man'' in the prompt ``a portrait of an old man''---and cases where they are not as ``cone hat'' and ``eating'' in the prompt ``A person wearing a cone hat is eating''~(see \cref{fig:inter_token_flow}). To achieve this, we use a large language model (LLM) as a judge. Specifically, we use GPT-4o and employ the following prompt:
\begin{quote}
``In Yes or No: in this prompt: \{input\_prompt\}, are \{item\_1\} and \{item\_2\} perceptually bound together?''
\end{quote}
We then filter out all cases where the lexical items are perceptually bound together and find that only 6.5\% instances exhibit unintentional leakage.

\subsection{Intended Item Evaluation}
\label{app:intended_pipeline}
For each lexical-item, we manually create two interpretations - one in the intended interpretation from the prompt, and another is a possible wrong interpretation of the word in other contexts. For example, given the prompt ``A standing zebra to the right of a city bus station'', the current interpretations would be ``the animal zebra'', while the incorrect interpretations would be ``A zebra crossing''. We then ask a VLM to evaluate the generated image, and asses if the first or the latter interpretations exists in the images.

To evaluate the model's capacity for contextual disambiguation, we manually define two interpretations for each lexical item in each prompt. The first is the intended semantic meaning derived from the prompt's context, and the second is an alternative interpretation, that is wrong in this context. For instance, in the prompt ``A standing zebra to the right of a city bus station'', the intended meaning of ``zebra is the animal'', whereas the wrong interpretation is a ``zebra crossing''. Subsequently, a VLM analyzes the generated image to determine if it depicts the intended interpretation or the wrong interpretation as we define it.

\subsection{Resources}
\label{app:resources}
Our computational experiments involved inference with four distinct text-to-image models: Flux-dev, Flux-schnell, sdxl-turbo, and Sana. The parameter sizes for these models are approximately 12 billion for Flux-dev and Flux-schnell, 3.1 billion for sdxl-turbo, and a range of 0.6 to 4.8 billion for the Sana models, with our experiments utilizing a 1.6 billion parameter version.  The total computational budget for these experiments is estimated to be approximately 480 GPU hours. The computing infrastructure consisted of a cluster of eight NVIDIA A100 GPUs. This configuration provided the necessary computational power for the large number of inference tasks performed.

\subsection{Use of AI Assistants.}
\label{app:ai}
We utilized AI assistants to support this research. For coding the experiments, we used Microsoft's Copilot and Anthropic's Claude 3; all generated code was manually reviewed and validated by us to ensure it aligned with our requirements. For the paper, Google's Gemini models (Pro and Flash) were used to improve the writing and clarity. We have carefully reviewed all content to ensure it accurately reflects our intentions.

\section{Data}
\label{app:data}

\textbf{DrawBench}~\citep{saharia2022photorealistic}: We include all categories except for ``misspelling'', ``rare words'', and ``text''.  Overall we extract 134 prompts from DrawBench.

In total, we obtain 1{,}053 prompts.

\begin{figure}[h]
  \centering
  \includegraphics[width=.9\linewidth]{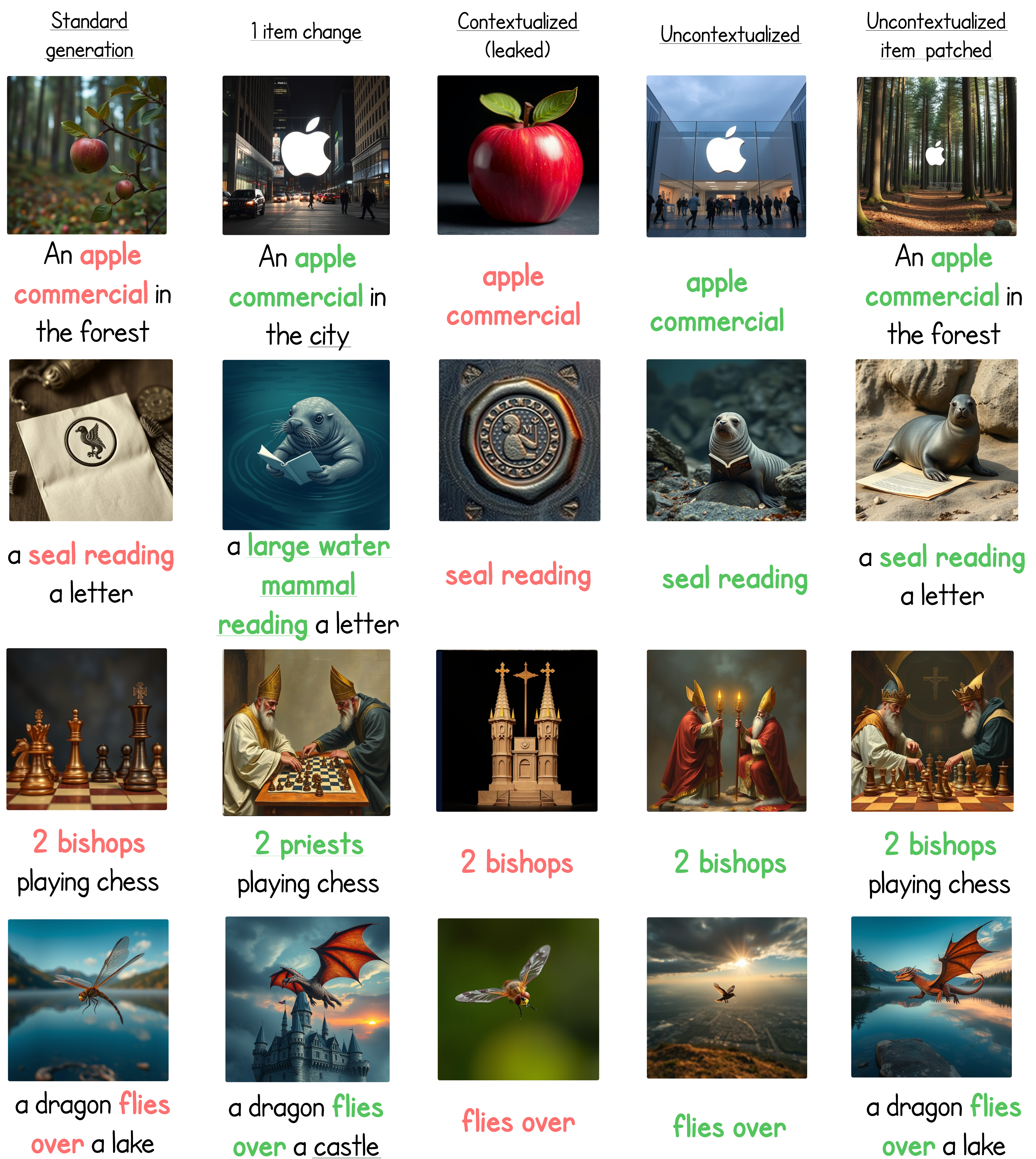}
  \caption{Examples from our semantic leakage method. \textbf{Left}: standard generation of leakage contained prompt. \textbf{Second}: generation using a one-lexical item change test as part of the dataset creation (a minimal substitution to verify that a slight lexical change yields a different image). \textbf{Third}: image from the contextual representation (misinterpreted item). \textbf{Fourth}: image from the uncontextualized representation (correct interpretation). \textbf{Right}: final generation after patching the correct, uncontextualized representation into the prompt.}
  \label{app:fig:additional_cleaned_examples}
\end{figure}

\paragraph{Extended dataset of leakage prompts.}
\label{app:leakage_prompts}
Our augmentation process incorporates two components. First, we generate variations of existing prompts from \cite{rassin-etal-2022-dalle} (e.g., modifying `a gentleman with a bow in the forest' to `a man wearing a bow in the jungle'). Second, we introduce novel prompts with potential semantic leakage. For these prompts, we apply a one-lexical item change test by generating an image from a similar prompt that substitutes the affected or leaked item with an alternative term (e.g., replacing `bishops' with `cardinals' or `checkers' in `chess in `2 bishops playing chess'). This test ensures that minimal lexical modifications do not alter the intended semantic meaning while producing a different image due to semantic leakage from another item in the prompt~(see the first two columns in \cref{app:fig:additional_cleaned_examples} for few visual examples). Together, these methods enrich the dataset and provide a robust framework for analyzing semantic leakage. The full list of prompts is available in our Git repository.\footnote{\href{https://github.com/tokeron/lens}{\texttt{https://github.com/tokeron/lens}}}

\section{Additional models}

\subsection{FLUX-Dev}
\label{app:flux_dev}
In addition to our primary experiments with FLUX, we repeated all analyses using the Flux-dev variant. The redundant versus representative token experiments yielded similar trends, with 55\% of tokens identified as representative and 45\% as non-representative—values closely matching those observed with FLUX. Likewise, our inter-item flow experiments confirmed that information flow occurred in 11\% of cases (and 3.1\% miss intended leakage), reinforcing the overall patterns reported in the main text. Notably, while the aggregate trends are consistent across models, the specific lexical items resolved can differ between FLUX-schnell and Flux-dev, indicating a potentially slightly different inner-lexicon~\citep{kaplan2025tokenswordsinnerlexicon}. These findings underscore the robustness of our approach while highlighting model-dependent nuances in token representation and information flow dynamics.

\subsection{SDXL-Turbo}
\label{app:SDXL-Turbo_comparison}

\begin{figure}[h]
  \centering
  \includegraphics[width=.9\linewidth]{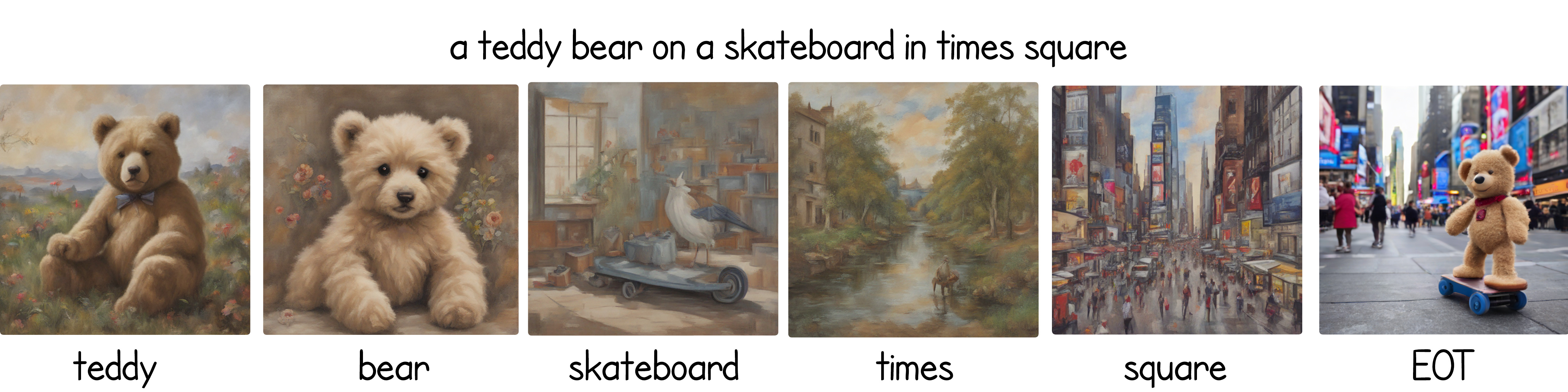}
        \caption{Images generated from individual subtokens in SDXL-Turbo. We find that, in many cases, the representation of an item is not clearly reflected in any of its subtokens—for example, in the case of the token ``skateboard.'' Another interesting observation is that the last token of a lexical item often carries its representation, as seen in the ``square'' token of ``times square.'' We also observe that the EOT token incorporates information from the full prompt.}
  \label{fig:sdxl_examples}
\end{figure}

Our analysis reveals that SDXL-Turbo, which uses the CLIP text encoder, behaves markedly differently from FLUX, which relies on the encoder in the encoder-decoder T5-XXL. In SDXL-Turbo, the text encoder is a causal language model, meaning each token's encoding is influenced only by its preceding tokens during the encoding process.

\begin{figure*}[h]
  \centering
  \includegraphics[width=1.\linewidth]{figures/sana_examplesV2.jpeg}
  \caption{Token-level image generation using Sana on the prompt \textit{``a black baseball hat with a flame decal on it''}. Representative information often resides in the last token of each item (e.g., ``hat'', ``decal''), consistent with Sana's causal encoding. Contextual influence is one-directional, with earlier tokens shaping later ones.}
  \label{fig:sana_examples}
\end{figure*}

We repeated our~\intra~experiments using SDXL-Turbo. Our first observation is that most generated images are either abstract or unrelated to the intended lexical items. According to our analysis, 55\% of lexical items in SDXL-Turbo lack any representative token (compared to just 11\% in FLUX). Moreover, when a representative token is present in CLIP, it is typically the final token of the lexical item (see \cref{fig:sdxl_examples}). This is aligned with the unidirectional encoding of the model.

Another phenomenon we observe—consistent with CLIP's training objective—is the unusually dominant role of the end-of-sequence (EOS) token. Images generated from the EOS token often encapsulate nearly the full semantic content of the prompt. In our evaluation, 62\% of EOS-generated images matched the prompt ( compared to 73\% when using the full prompt). We believe this also causes our intervention method to be less effective, since when we interpret a single token, we patch all other tokens, including the EOT token—which usually contains a lot of information—with tokens derived from an empty prompt (see \cref{fig:sdxl_examples}).

\subsection{Sana (Gemma-based encoder)}
\label{app:sana_comparison}

We also evaluate our methodology on the Sana model~\citep{xie2024sana}, which uses a Gemma-based autoregressive language model as the text encoder. These results help validate the generality of our findings across architectures with differing encoding strategies.

For \textit{in-item information flow}, we observe that Sana produces fewer multi-token lexical items due to its larger vocabulary relative to T5 and CLIP. In cases where items do consist of multiple tokens, we find that only the \textit{last} token typically acts as a representative token—a behavior aligned with the unidirectional nature of autoregressive encoders.

In our \textit{cross-item information flow} analysis, we find that 17.45\% of item pairs in Sana exhibit contextual information flow, compared to 10.45\% in FLUX. This increase likely stems from the one-sided (forward-only) nature of Sana's encoding. For example, in the prompt \textit{``a black baseball hat with a flame decal on it''}, we find contextual influence such as ``black baseball'' affecting ``hat'' and ``flame'' affecting ``decal'', but not the reverse. See~\cref{fig:sana_examples} for illustrations.

Overall, while Sana shows a slightly higher rate of contextual influence than FLUX, it preserves many of the key structural insights found in our primary analysis—particularly the sparsity and location of representative tokens. Unlike CLIP, which encodes substantial information in the EOS token, Sana lacks such artifacts, suggesting that EOS-related effects are not fundamental to autoregressive models more generally.

\section{Redundant Token Classifier Comparison}
\label{appendix:classifiers}

To identify redundant tokens efficiently, we evaluated several lightweight classifiers operating directly on token embeddings from the text encoder, without requiring any image generation. All classifiers were trained on the same dataset of token-level annotations produced by the patching technique described in \secref{Intra-concepts}, using an 80/20 train-validation split.

We compared three classifiers: a k-nearest neighbors classifier (k-NN, $k=5$, Euclidean distance), logistic regression (L2 penalty, balanced class weights), and a single linear layer trained with binary cross-entropy loss for 25 epochs. All classifiers were evaluated using accuracy, precision, recall, and F1-score, with results reported in~\cref{tab:classifier_full}.

\begin{table}[h]
\centering
\resizebox{\linewidth}{!}{%
\begin{tabular}{lcccc}
\toprule
Classifier & Accuracy & Precision & Recall & F1-score \\
\midrule
k-NN ($k=5$)          & 0.82 & 0.92 & 0.74 & 0.82 \\
Logistic Regression   & 0.80 & 0.90 & 0.81 & 0.85 \\
Linear Layer (NN)     & 0.83 & 0.90 & 0.86 & 0.88 \\
\bottomrule
\end{tabular}%
}
\caption{Comparison of lightweight classifiers for predicting token redundancy from encoder hidden states.}
\label{tab:classifier_full}
\end{table}

All three classifiers achieve precision above 90\%, confirming that representative tokens are reliably identifiable from encoder hidden states across a range of approaches. The linear layer achieves the best overall balance, with the highest recall and F1-score, while maintaining competitive precision. The k-NN classifier achieves the highest precision but at the cost of lower recall, meaning it is more conservative in flagging tokens as redundant. We use the linear layer as our default classifier in the main paper.

\section{Inter-Item Information Flow Framework}
\label{app:intervention_inter}

To assess whether one lexical item encodes information about others in the same prompt, we conduct the following experiment.

Given a prompt, we isolate each lexical item one at a time. First, we encode the full prompt using the text encoder. Then, for a given lexical item, we apply our patching method~(\cref{sec:method}) to mask the representations of all other tokens—leaving only the contextualized representation of the target item intact. Formally, for a lexical item with token indices \( S \subset \{1, \ldots, N\} \), we construct a patched sequence \( \tilde{t}_1, \ldots, \tilde{t}_N \) in which only the tokens in \( S \) retain their original contextualized representations. We then generate an image from this modified sequence, capturing what information is encoded in the selected item’s contextualized form.

We repeat this process for the same item in an uncontextualized setting: we encode and generate an image using only that lexical item in isolation, without the rest of the prompt. This allows us to distinguish between information inherently present in the item’s encoding and information introduced by context.

To measure influence between items, we use Qwen2-VL to check whether a second item \( y \) appears in the image generated from a first item \( x \), both in the contextualized and uncontextualized versions. Influence is considered \textit{True} if \( y \) appears in the image generated from contextualized \( x \), but not from uncontextualized \( x \). This comparison ensures that the observed presence of \( y \) is attributable to contextual information flow during encoding, rather than coincidental co-occurrence or inherent correlation.

This setup enables us to detect breaches of lexical segmentation—i.e., cases where one lexical item encodes visual information belonging to another—quantifying interdependence across items within the prompt.

\begin{figure*}[t]
  \centering
  \includegraphics[width=\linewidth]{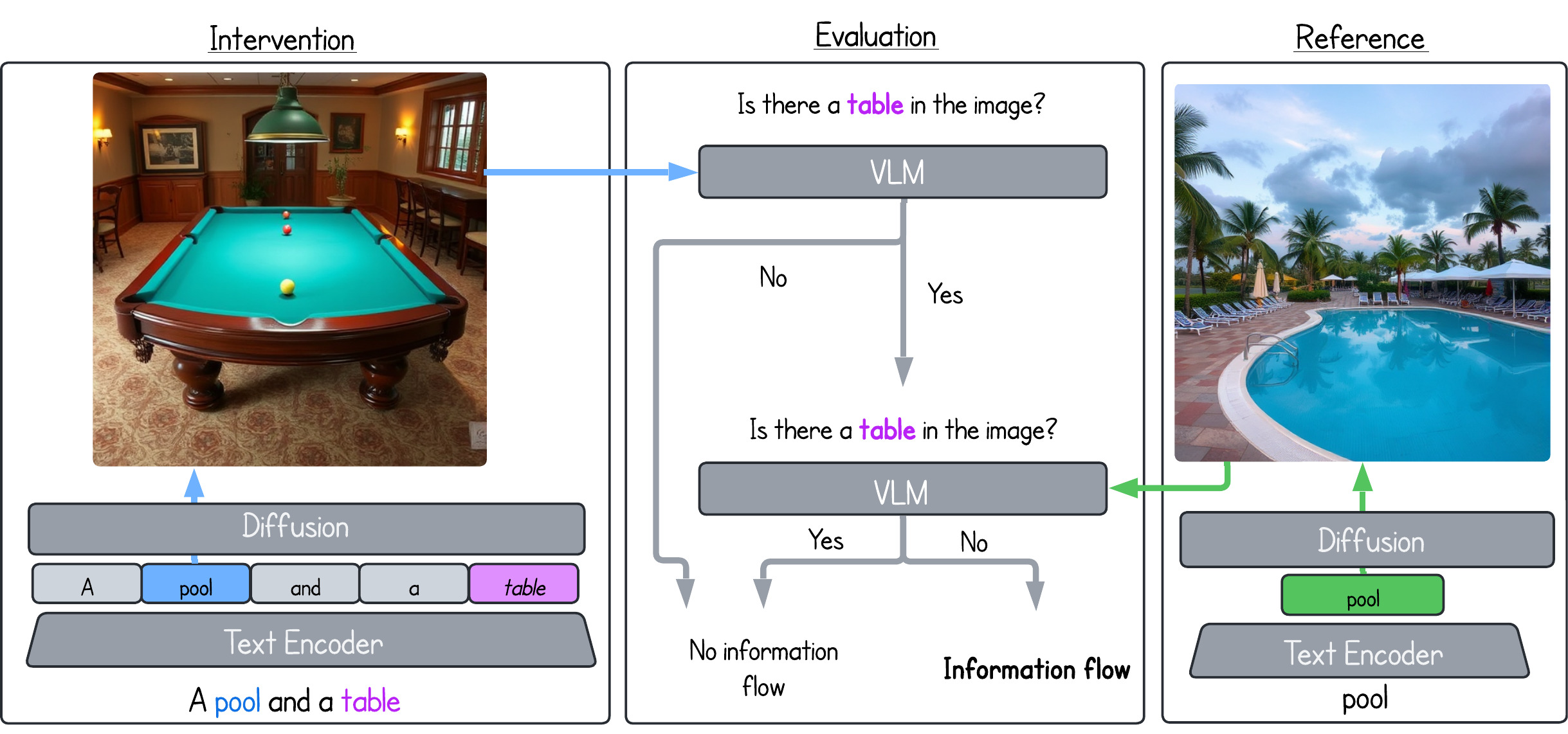}
    \caption{\textbf{Evaluating inter-item information flow:} Our proposed framework to interpret the information flow between lexical items in the prompt. For each lexical item, we generate an image from its contextual representations~(left), and from it's uncontextualized representation (right), and analyze the generated images using a VLM~(middle). In this example, we interpret the item \textcolor{poolblue}{``pool''} and assess whether it is influenced by the item \textcolor{tablepurple}{``table''}. To do so, we ask a VLM whether the image generated from the token \textcolor{poolblue}{``pool''} contains a \textcolor{tablepurple}{``table''}. To ensure this is the result of information flow, and not a natural correlation between a pool and tables, we generate the item \textcolor{greenpool}{``pool''} without context (right-hand side), and verify that \textcolor{tablepurple}{``table''} is not present in this image. If this is the case, we conclude that there was information flow from \textcolor{tablepurple}{``table''} to \textcolor{poolblue}{``pool''}.    }
  \label{fig:framework_inter}
\end{figure*}

See \cref{fig:framework_inter} for an illustration of the procedure.
\begin{figure}[h]
  \centering
  \includegraphics[width=.9\linewidth]{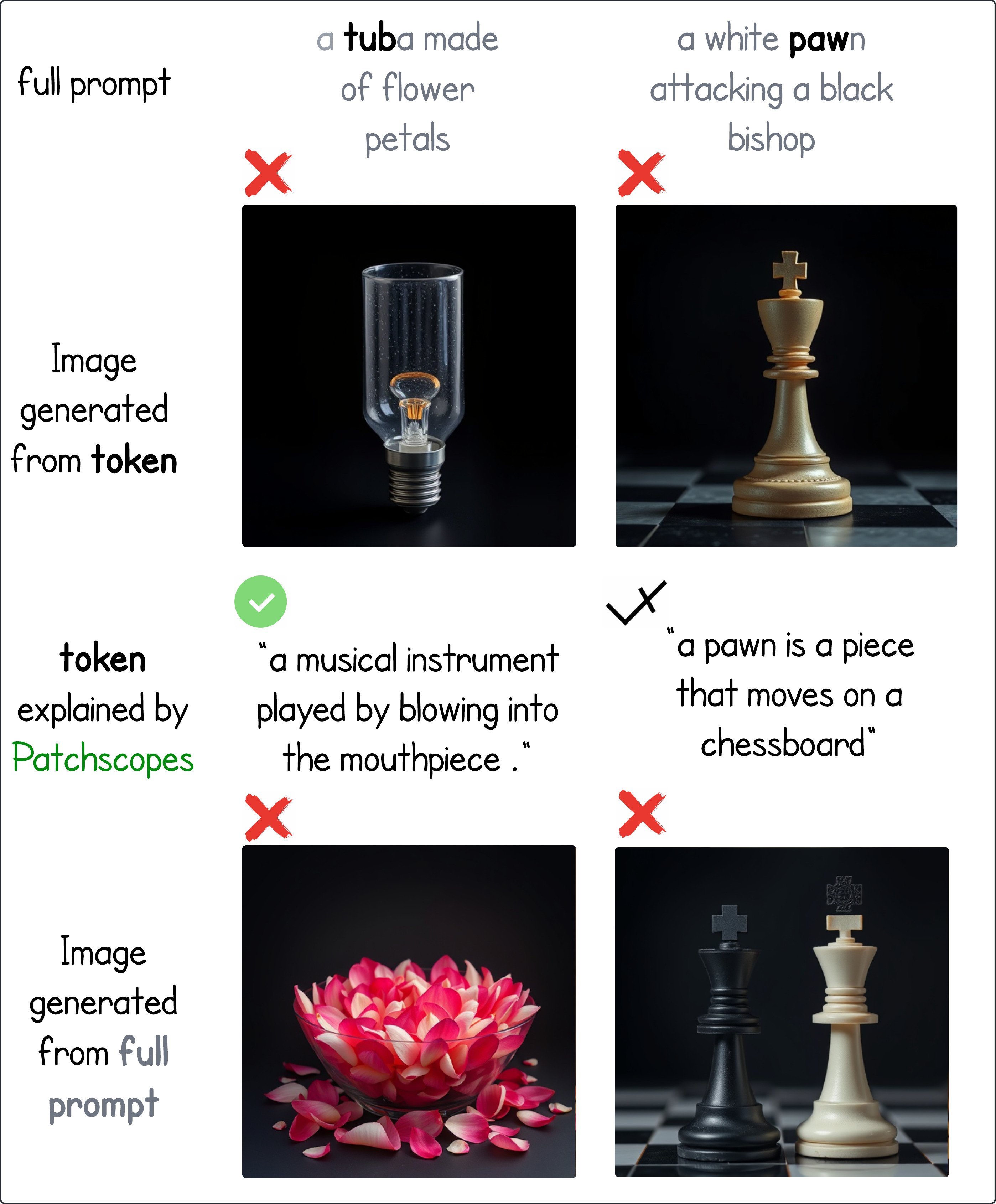}
        \caption{
    Comparing Patchscopes to image generation with token-level patching in cases of~\neglect. We assess whether a token’s concept is preserved by comparing images generated from its contextual representation, Patchscopes’ textual interpretation, and the full prompt image. \textbf{Left:} ``tub'' (from ``tuba'') is correctly described by Patchscopes but fails to ground visually. \textbf{Right:} ``paw'' (from ``pawn'') has missing details by both interpretations, suggesting a gap in the encoder’s conceptual knowledge.}
  \label{fig:textual_image_miss_alignment}
\end{figure}

\begin{figure}[t]
  \centering
  \includegraphics[width=0.6\linewidth]{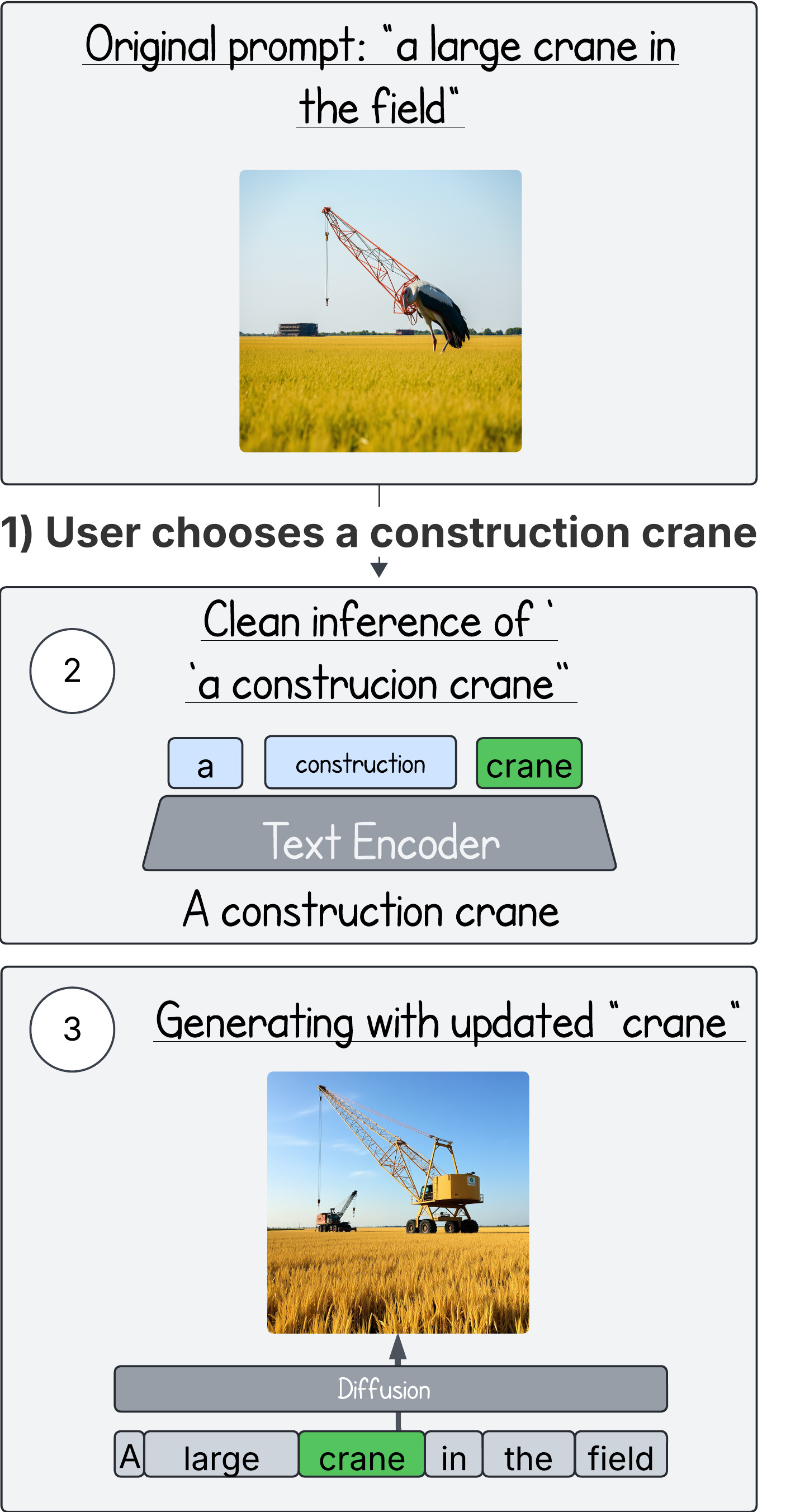}
  \caption{\textbf{Resolving polysemous words using patching.} 
  Example: in the prompt \textit{``a crane in the field''}, the word \textbf{crane} may refer either to a construction machine or to a bird. 
  (1) The user specifies the intended meaning. 
  (2) The prompt is encoded with this interpretation. 
  (3) We patch the encoded representation and generate an image, producing a result consistent with the user’s choice.}
  \label{fig:patching_application}
\end{figure}

\section{Analysis of Negligence: Understanding Omitted Items}
\label{app:neglect}

We analyze cases where a lexical item is completely absent from the generated image, a phenomenon we refer to as \neglect~\citep{chang-etal-2024-repairing,chefer2023attend}. 
As shown in \cref{fig:rep-token-dist}, when a lexical item has no representative tokens, it is also missing from an image generated using all of its tokens in 88\% of cases. 
This strong correlation suggests a direct connection between the absence of representative tokens and the failure to visually render certain concepts.

\paragraph*{Hypotheses}
To understand why this occurs, we consider two potential explanations.  
First, the text encoder may fail to capture the concept adequately, resulting in a weak or missing internal representation.  
Second, the concept may be well-encoded in text, but the diffusion model itself may lack familiarity with the concept’s visual appearance, preventing it from rendering the item correctly.

\paragraph*{Methodology}
To distinguish between these two possibilities, we analyze the text encoder independently of the diffusion model using Patchscopes~\citep{Ghandeharioun2024PatchscopesAU}.  
Patchscopes decodes token-level representations into natural language descriptions through the full encoder-decoder architecture of T5.  
For each neglected item, we patch its encoding into the template:

\begin{quote}
\texttt{describe <item>}
\end{quote}

We then evaluate whether the resulting description accurately represents the intended concept using GPT-4o~\citep{openai2024gpt4technicalreport}.  
Details of the GPT-4o evaluation prompt are provided in Appendix~\ref{app:gpt-ps}.  
This procedure is applied to all neglected lexical items identified in Section~\ref{sec:how_is_info_distributed}.

\paragraph*{Results}
Our analysis shows that in 67\% of the neglected cases, Patchscopes returns accurate descriptions.  
For instance, as illustrated in \cref{fig:textual_image_miss_alignment}, the token sequence for ``tuba'' yields the description:  
\textit{``a musical instrument played by blowing into the mouthpiece.''}  

These results suggest that for the majority of neglected items, the text encoder correctly encodes the concept, and the failure likely arises within the diffusion model’s visual grounding process.  
In the remaining 33\% of cases, the Patchscopes outputs are incomplete or incorrect, indicating that the text encoder itself lacks a robust representation of the item.

\paragraph*{Interpretation}
Taken together, these findings point to two distinct sources of item-level negligence:
(1) representational gaps in the text encoder, where the concept is poorly encoded or missing entirely; and  
(2) visual grounding failures in the diffusion model, where the concept is well-encoded but cannot be visually rendered.  

The first failure mode aligns with prior work showing that strengthening textual descriptions can improve factual generation~\citep{huang-etal-2025-t2i}.  
The second highlights a limitation of the diffusion model’s training distribution and its capacity to map textual meaning to visual form.

Additional qualitative examples of these two cases are provided in Figure~\ref{fig:textual_image_miss_alignment}, where we show how different concepts behave under our Patchscopes-based evaluation.

\section{Resolving Polysemous Words via Patching}
\label{app:polysemous}

Polysemous words such as ``bat,'' ``bow,'' or ``crane'' can correspond to multiple meanings, and current T2I models often default to the most common interpretation or fail to disambiguate. In some cases, users may even prefer a surprising reading (e.g., \textit{flying baseball bats on top of a baseball stadium}), or the context alone may not suffice (e.g., \textit{a gentleman wearing a bow in the forest}). 

To address this, we extend our patching technique to allow explicit user control over interpretation. Given a prompt and a user-specified meaning, we replace the ambiguous token with a clean, context-independent representation of the intended concept, while leaving the surrounding context unchanged. 

We demonstrate this approach over several ambiguous words, each paired with two distinct interpretations. Ambiguous prompts were generated using GPT-4o, which also produced disambiguated representations of each meaning. For every prompt, we generated images and used a vision-language model (VLM) to assess whether the output matched the user-specified interpretation. Our method achieved a 95\% success rate in aligning images with the chosen interpretation. 

As shown in Figure~\ref{fig:patching_application}, this patching-based approach enables precise and reliable resolution of lexical ambiguity, offering improved control for both faithful image generation and creative use cases.

\section{Semantic Leakage in Closed SOTA Models}
\label{appendix:closed_models}

To assess whether semantic leakage generalizes beyond the open models studied in the main paper, we conducted additional experiments on a closed SOTA text-to-image model (Nano-Banana~\citep{geminiteam2025geminifamilyhighlycapable}). Since internal token representations are inaccessible in such models, we adapted our evaluation protocol to rely solely on generated images.

We use the same 110 prompts from our semantic leakage evaluation (\secref{inter_item}). For each prompt containing a polysemous item, we construct a paired prompt by substituting the contextual cue with a neutral alternative that should not affect the intended interpretation of the polysemous word, but would produce a different image if semantic leakage occurs. For example, for the prompt \textit{"bats flying above a baseball stadium"}, where \textit{bat} should be interpreted as the flying animal, we construct the paired prompt \textit{"bats flying above a football stadium"}. Both prompts are then passed to the closed model, and a VLM evaluates whether the intended interpretation of the polysemous item appears in each generated image. We mark a case as semantic leakage when the neutralized version produces the intended interpretation but the original does not---indicating that the contextual cue distorted the item's meaning.

The closed model exhibits semantic leakage in 26\% of prompts, substantially lower than FLUX-schnell (94\%) and FLUX-dev (79\%), yet far from negligible. Leakage is also clearly directional: when the polysemous item appears \textit{after} its contextual cue in the prompt, the leakage rate rises to 42\%, compared to lower rates when the item precedes the cue. This asymmetry is consistent with the causal encoding effects observed in our open-model analysis, where earlier tokens disproportionately influence the representations of later ones.

These results demonstrate that encoder-level semantic leakage is not an artifact of specific open architectures or tokenization schemes, but a general property of current T2I systems. The reduced rate in the closed model may reflect architectural improvements or larger-scale training, but the persistence of directional leakage suggests the underlying mechanism remains. Notably, since we cannot apply our patching-based mitigation to closed models, these findings further motivate developing encoder-aware solutions that can generalize across architectures.

\end{appendices}

\end{document}